\documentclass[letterpaper]{article} % DO NOT CHANGE THIS
\usepackage{aaai25}  % DO NOT CHANGE THIS
\usepackage{times}  % DO NOT CHANGE THIS
\usepackage{helvet}  % DO NOT CHANGE THIS
\usepackage{courier}  % DO NOT CHANGE THIS
\usepackage[hyphens]{url}  % DO NOT CHANGE THIS
\usepackage{graphicx} % DO NOT CHANGE THIS
\usepackage{natbib}  % DO NOT CHANGE THIS AND DO NOT ADD ANY OPTIONS TO IT
\usepackage{caption} % DO NOT CHANGE THIS AND DO NOT ADD ANY OPTIONS TO IT
\usepackage{algorithm}
\usepackage{algorithmic}

\usepackage{newfloat}
\usepackage{listings}
\DeclareCaptionStyle{ruled}{labelfont=normalfont,labelsep=colon,strut=off} % DO NOT CHANGE THIS
\floatstyle{ruled}
\newfloat{listing}{tb}{lst}{}
\floatname{listing}{Listing}
\usepackage{multirow}
\usepackage{amsmath}
\usepackage{amssymb}
\usepackage{mathtools}
\usepackage{amsthm}
\usepackage{thmtools}
\usepackage{thm-restate}
\usepackage{amsfonts}
\usepackage{bbding}
\usepackage{algorithm}
\usepackage{algorithmic}
\usepackage{cleveref}
\usepackage{caption}
\usepackage{subfigure}
\usepackage{graphicx}
\usepackage{booktabs} 
\usepackage[dvipsnames]{xcolor}

\newcommand{\mbf}{\mathbf}
\newcommand{\mca}{\mathcal}
\newcommand{\mbb}{\mathbb}
\newcommand{\mrm}{\mathrm}
\newcommand{\moprm}[1]{\mathop{\mathrm{#1}}}
\newcommand{\tit}{\textit}
\newcommand{\tbf}{\textbf}
\newcommand{\ttt}{\texttt}
\newcommand{\udl}{\underline}
\theoremstyle{plain}
\newtheorem{theorem}{Theorem}[section]

\theoremstyle{definition}
\newtheorem{definition}[theorem]{Definition}

\theoremstyle{remark}

\title{Single-View Graph Contrastive Learning with Soft Neighborhood Awareness}
\author {
    Qingqiang Sun\textsuperscript{\rm 1},
    Chaoqi Chen\textsuperscript{\rm 2},
    Ziyue Qiao\textsuperscript{\rm 1}\thanks{Corresponding Author},
    Xubin Zheng\textsuperscript{\rm 1},
    Kai Wang\textsuperscript{\rm 3}\thanks{Corresponding Author},
}
\affiliations {
    \textsuperscript{\rm 1}Great Bay University,
    \textsuperscript{\rm 2}Shenzhen University,
    \textsuperscript{\rm 3}Central South University\\
    \{qqsun, zyqiao, xbzheng\}@gbu.edu.cn, cqchen1994@gmail.com, kaiwang@csu.edu.cn
}
\usepackage{bibentry}
\begin{document}

\maketitle

\begin{abstract} \label{abs}
Most graph contrastive learning (GCL) methods heavily rely on cross-view contrast, thus facing several concomitant challenges, such as the complexity of designing effective augmentations, the potential for information loss between views, and increased computational costs. To mitigate reliance on cross-view contrasts, we propose \ttt{SIGNA}, a novel single-view graph contrastive learning framework. Regarding the inconsistency between structural connection and semantic similarity of neighborhoods, we resort to soft neighborhood awareness for GCL. Specifically, we leverage dropout to obtain structurally-related yet randomly-noised embedding pairs for neighbors, which serve as potential positive samples. At each epoch, the role of partial neighbors is switched from positive to negative, leading to probabilistic neighborhood contrastive learning effect. Furthermore, we propose a normalized Jensen-Shannon divergence estimator for a better effect of contrastive learning. Surprisingly, experiments on diverse node-level tasks demonstrate that our simple single-view GCL framework consistently outperforms existing methods by margins of up to $21.74\%$ (\tit{PPI}). In particular, with soft neighborhood awareness, \ttt{SIGNA} can adopt MLPs instead of complicated GCNs as the encoder to generate representations in transductive learning tasks, thus speeding up its inference process by $109\times$ to $331\times$. The source code is available at \url{https://github.com/sunisfighting/SIGNA}.
\end{abstract}

%     \link{Code}{https://anonymous.4open.science/r/SIGNA}
%     % \link{Datasets}{https://aaai.org/example/datasets}
%     % \link{Extended version}{https://aaai.org/example/extended-version}

\section{Introduction}
\label{sec: intro}

% Over the past few years, deep learning on graphs has achieved unprecedented success in a lot of tasks~\citep{wu2021survey, liu2022survey, xie2022survey}. 
% % such as node classification~\citep{kipf2016GCN, hamilton2017graphsage}, link prediction~\cite{kipf2016GAE, zhang2018link}, graph classification~\citep{zhang2018graphclass, errica2019graphclass}. 
% However, most supervised and semi-supervised methods heavily rely on the quality and quantity of label data. In many real-world scenarios, obtaining such data is costly and challenging, thereby constraining the broader applicability of these techniques. Addressing this limitation, 
Over the past few years, Self-Supervised Learning (SSL) of representations has emerged as a promising and popular research topic since it does not rely on the quantity and quality of labels~\citep{wu2021survey, liu2022survey, xie2022survey}. As one of the most competitive SSL paradigms, Contrastive Learning (CL) has made its mark in computer vision~\citep{he2020MOCO, chen2020SIMCLR}, natural language processing~\citep{tian2020CMC}, and graph domains~\citep{velickovic2019DGI, zhu2020GRACE, peng2020GMI, sun2024interdependence}. 

\begin{figure}
  \centering   \includegraphics[width=0.98\linewidth]{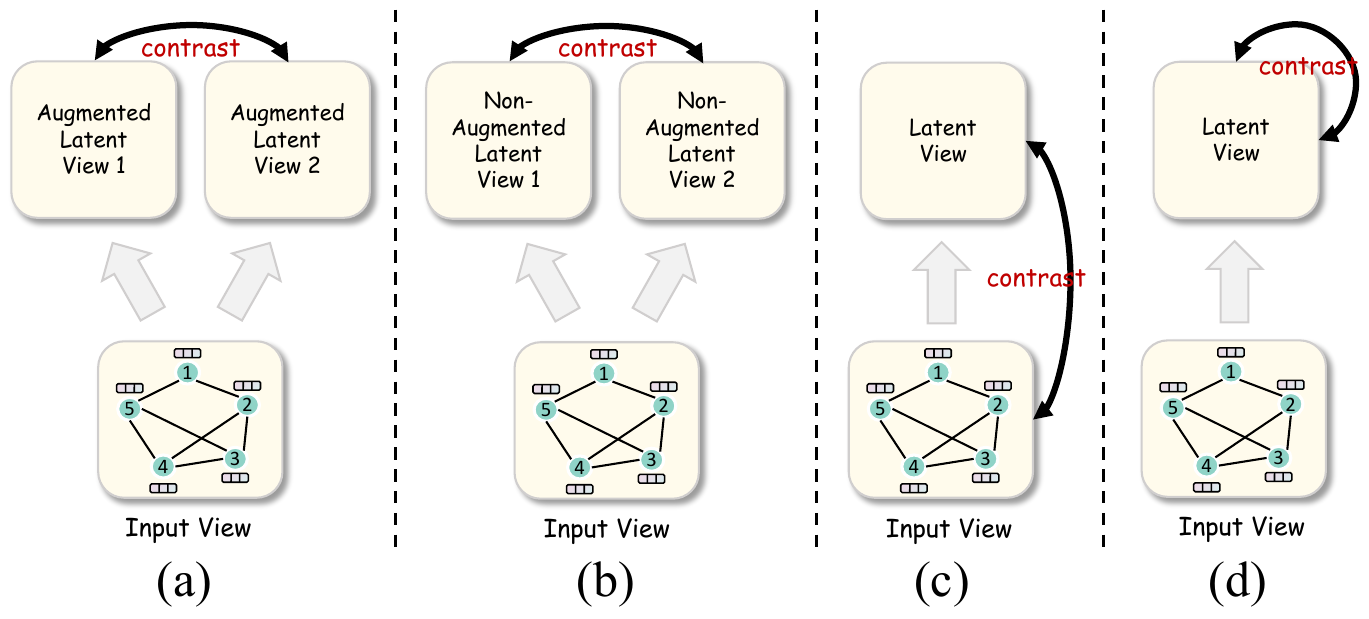}
   \caption{(a) Augmentation based cross-view contrast; (b) Non-augmentation based cross-view contrast; (c) Input-latent cross-view contrast; (d) Single-view contrast (ours).}
   \label{fig: contrast_view}
\end{figure}

The core spirit of CL is to pull together representations of related instances (positive samples) and push apart representations of unrelated pairs (negative samples). Similar to CL in computer vision and natural language processing domains, Graph Contrastive Learning (GCL) methods typically rely on cross-view contrasts, \tit{i.e.}, each positive or negative pair consists of two latent embeddings from different views. These strategies include augmentation-based dual-branch methods~\citep{zhu2020GRACE, you2020GraphCL, shen2023NCLA}, non-augmentation-based dual-branch methods~\citep{thakoor2021BGRL, mo2022SUGRL}, and input-latent single-branch methods~\citep{peng2020GMI}, which are summarized in Fig.~\ref{fig: contrast_view} and discussed in detail in the related work. 

Despite the significant progress achieved by cross-view contrastive learning methods on graph data, these approaches face several challenges. Firstly, designing effective augmentation techniques often requires substantial manual effort and fine-tuning, leading to increased implementation complexity. Secondly, cross-view contrast may result in information loss or inconsistencies between views, which can negatively impact model performance. Lastly, the computational cost of cross-view methods is typically high, especially for those dual-branch ones, making them less scalable.

To circumvent these challenges, this paper seeks to explore a simpler yet little-studied alternative: single-view contrastive learning. The most pivotal point is how to obtain intra-view positive and negative samples. Note that nodes in the graph are intricately correlated with each other rather than completely independent~\citep{dang2021nearest, wang2022ada, yu2024contextually, wang2023towards, YiqiWang2024VLDB}. Therefore, a natural potential solution within this framework is to conduct contrast according to the topological relationship between nodes, \tit{i.e.}, pull embeddings of neighbors together and push those of non-neighbors away. However, such a trivial solution may be prone to overfitting  topological information and thus results in suboptimal performance. A typical example is GAE~\citep{kipf2016GAE}, which aims to reconstruct the graph structure as much as possible but can only achieve less competitive performance than other self-supervised methods that ignore topological information, such as GRACE~\citep{zhu2020GRACE}. In such a context, two questions are naturally raised: \tit{(i) What role should neighborhoods play in node-level GCL? (ii) Can neighborhood awareness help single-view contrastive learning surpass cross-view contrastive learning?}

Regarding these two questions, we first investigate the homophily nature across diverse real-world datasets from both global and local perspectives. Statistical results indicate that there is an obvious inconsistency between structural connection and semantic similarity. Inspired by this, we propose \tbf{S}ingle-v\tbf{I}ew \tbf{G}raph contrastive learning with soft \tbf{N}eighborhood \tbf{A}wareness (\ttt{SIGNA}). Instead of relying on augmentations or other cross-view contrastive techniques to generate contrastive pairs,
% ~\citep{he2020MOCO, chen2020SIMCLR, zhu2020GRACE, zhu2021GCA, thakoor2021BGRL}, 
\ttt{SIGNA} adopts soft neighborhood awareness to realize single-view contrast. Specifically, we first use dropout to obtain randomly-perturbed embeddings, which implicitly provides more diverse embedding combinations for contrast. At each epoch, the role of neighbors is allowed to switch from positive to negative rather than being consistently designated as positive, thus leading to probabilistic neighborhood contrastive learning. Furthermore, we propose a normalized Jensen-Shannon divergence (Norm-JSD) estimator, which combines the advantages of both JSD and InfoNCE, resulting in a better contrastive effect.

We evaluate \ttt{SIGNA} on three kinds of node-level tasks, including transductive node classification, inductive node classification, and node clustering. \ttt{SIGNA} consistently outperforms existing methods across all tasks, achieving performance gains of up to $21.74\%$ (\tit{PPI}). In particular, thanks to reasonable neighborhood awareness, \ttt{SIGNA} is enabled to use a simple Multilayer Perceptron (MLP) as the encoder in transductive learning tasks, which is $109\times$ to $331\times$ faster than a GCN-based encoder with the same settings in terms of the inference time. The distribution analysis and visualization of learned representations showcase the superiority of \ttt{SIGNA} in striking a better balance between intra-class aggregation and inter-class separation, compared with representative neighborhood-aware GCL methods. In summary, {our contributions can be highlighted as follows:}

$\bullet$ We study the role of neighborhoods by comprehensively investigating their homophily nature on real-world datasets from both global and local perspectives, thus offering nuanced insights on soft neighborhood awareness.  

$\bullet$ We propose a simple yet effective single-view GCL framework, \ttt{SIGNA}, which alleviates the heavy reliance on traditional cross-view contrastive learning with the help of our soft neighborhood awareness strategy.

$\bullet$ Extensive experiments demonstrate that \ttt{SIGNA} outperforms existing methods across various node-level tasks, showcasing the feasibility of single-view GCL and the rationality of soft neighborhood awareness.

\section{Related Works}
\label{sec: related work}
To position our contributions in the literature, we briefly review related works here.

\tbf{GCL Paradigms.} Existing GCL methods mainly rely on cross-view contrast, which can be further grouped into three classes, \tit{i.e.}, augmentation based cross-view contrast, non-augmentation based cross-view contrast, and input-latent cross-view contrast, as illustrated in Fig.~\ref{fig: contrast_view}. (a)-(c). (a) \tit{Augmentation based cross-view contrast} is represented by DGI~\citep{velickovic2019DGI}, MVGRL~\citep{hassani2020MVGRL}, GRACE~\citep{zhu2020GRACE}, GraphCL~\citep{you2020GraphCL}, GCC~\citep{qiu2020gcc}, and MERIT~\citep{jin2021merit}, NCLA~\citep{shen2023NCLA}, etc. Augmented views from the same instance are pulled closer while those views from distinct instances are pushed away. (b) Instead of augmenting input data, \tit{non-augmentation based cross-view contrast} resorts to using two discrepant encoders or directly perturbing latent embeddings to obtain contrastive pairs, such as SUGRL~\citep{mo2022SUGRL}, AFGRL~\citep{lee2022AFGRL}, SimGRACE~\citep{xia2022simgrace}, SimGCL~\citep{yu2022AugNecessary}, and COSTA~\citep{zhang2022costa}. (c) \tit{Input-latent cross-view contrast} is represented by GMI~\citep{peng2020GMI}, which aims to maximize mutual information between the input graph and latent embeddings. 

In this paper, we aim to study the simpler single-view GCL, which has seldom been studied in the literature.

% \tbf{Augmentation-invariant GCL.} Inspired by the remarkable success achieved in vision domain~\citep{wu2018instdist, chen2020SIMCLR, he2020MOCO}, typical GCL methods adopt augmentation-invariant paradigms. Represented by DGI~\citep{velickovic2019DGI}, MVGRL~\citep{hassani2020MVGRL}, GRACE~\citep{zhu2020GRACE}, GraphCL~\citep{you2020GraphCL}, GCC~\citep{qiu2020gcc}, and MERIT~\citep{jin2021merit}, augmented views from the same instance are pulled closer while those views from distinct instances are pushed away. Since optimal augmentation hyperparameters are sensitive to datasets and domains, some works resort to adaptive or dynamic augmentation techniques, such as GCA~\citep{zhu2021GCA} and JOAO~\citep{you2021autoCL}. Instead of augmenting input data, some works generate contrastive pairs via perturbing embeddings (SimGCL~\citep{yu2022AugNecessary}, COSTA~\citep{zhang2022costa}) or encoders (SimGRACE~\citep{xia2022simgrace}). However, solely relying on augmentations to obtain positives makes them incompetent to capture desirable semantics between neighbors. 

\tbf{Neighborhood-aware GCL.} Regarding the limitations of augmentation-invariant GCL, concurrent studies turn to neighborhood-aware techniques to retain semantics. For example, GMI~\citep{peng2020GMI} maximizes the mutual information across both feature and edge representations between the input and output spaces. GraphCL-NS~\citep{hafidi2022graphcl-ns} makes use of graph structure to sample negatives, \tit{i.e.}, from $l$-th order neighbors of the anchor node. Based on NT-Xent~\citep{zhu2020GRACE}, NCLA~\citep{shen2023NCLA} adopts a neighbor contrastive loss that regards both intra-view and inter-view neighbors as positives. Without generating augmented views, Local-GCL~\citep{zhang2022localized} fabricates positive samples for each node using first-order neighbors. Instead of binary contrastive justification, GSCL~\citep{ning2022GSCL} uses fine-grained contrastive justification according to the hop distance of neighborhoods. AFGRL~\citep{lee2022AFGRL} obtains local positives by jointly considering the adjacency matrix and k-NNs in the embedding space. SUGRL~\citep{mo2022SUGRL} employs both GCN and MLP encoders to obtain contrastive pairs. For each anchor node, its positive counterparts are constructed with its GCN output and the MLP outputs of its neighbors. 
% Although neighborhood-aware techniques favor structural information, its rationality in GCL remains unexplored. 

Unlike existing works that fully trust neighborhoods to capture semantic information for downstream tasks, we resort to soft neighborhood awareness to prevent the encoder from overfitting uncertain signals. 

\section{Rethinking the Role of Neighborhoods}

To implement effective sigle-view contrastive learning, it is of great significance to understand \tit{what role neighbours should play in graph contrastive learning}. Thus, we first empirically investigate the homophily nature of neighborhoods.

Let $\mca{G}=(\mca{V},\mca{E})$ be an unweighted graph with a node set $\mca{V}$ and an edge set $\mca{E}$. We denote the feature matrix and adjacency matrix by $\mbf{X}=\{\mbf{x}_i\}_{i=1}^{|\mca{V}|}\in \mbb{R}^{|\mca{V}|\times F}$ and $\mbf{A}\in\{0,1\}^{|\mca{V}|\times|\mca{V}|}$, respectively. The one-hot label matrix is denoted by $\mbf{Y}=\{\mbf{y}_i\}_{i=1}^{|\mca{V}|}\in\mbb{R}^{|\mca{V}|\times c}$. Following \citep{zhu2020beyondHomo}, the definition of global homophily ratio is given by:

\begin{definition}[Global Homophily Ratio]
Given a graph $\mca{G}=(\mca{V},\mca{E})$, its
global homophily ratio is defined as the probability that two connected nodes share the same label:
\begin{equation}
\mca{H}_{global}=\frac{1}{|\mca{E}|}\sum\nolimits_{u,v\in\mca{V}}\mathbb{I}[(u,v)\in\mca{E}]\cdot\mathbb{I}[\mbf{y}_u=\mbf{y}_v].
\end{equation}    
\end{definition}
Furthermore, we define local homophily count and local homophily ratio as well:

\begin{definition}[Local Homophily Count]
Given a node $u\in\mca{V}$ and its one-hop neighbors $\mca{N}_u$, its local homophily count is defined as the number of neighbors with the same label:
\begin{equation}
\mca{H}_{local}^{\#}(u)= \sum\nolimits_{v\in\mca{N}_u}\mathbb{I}[\mbf{y}_u=\mbf{y}_v].     
\end{equation}    
\end{definition}

\begin{definition}[Local Homophily Ratio]
Given a node $u\in\mca{V}$ and its one-hop neighbors $\mca{N}_u$, its local homophily ratio is defined as the probability that its neighbors share the same label with it:
\begin{equation}
\mca{H}_{local}(u)= \frac{1}{|\mca{N}_u|}\sum\nolimits_{v\in\mca{N}_u}\mathbb{I}[\mbf{y}_u=\mbf{y}_v].     
\end{equation}
\end{definition}

\begin{figure}[!t]
    \centering
    % First image (approximately half the line width)
    \begin{minipage}{0.45\textwidth}
    \includegraphics[width=\linewidth]{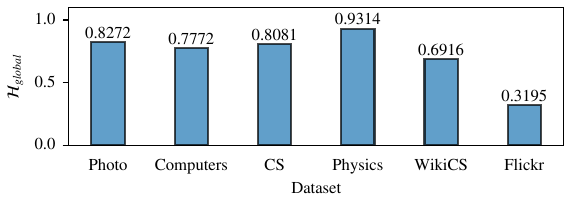}
    \end{minipage}
    \begin{minipage}{0.22\textwidth}
    \includegraphics[width=\linewidth]{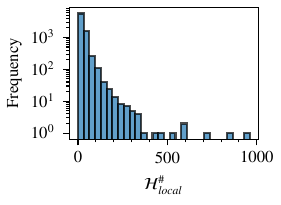}
    \end{minipage}
    % \hfill % Adds horizontal space between the minipages
    % Third image (approximately a quarter of the line width)
    \begin{minipage}{0.22\textwidth}    \includegraphics[width=\linewidth]{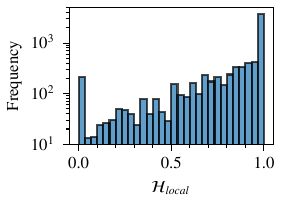}
    \end{minipage}
    \caption{Global and local homophily statistics. \tit{Top}: global homophily ratios on different datasets. \tit{Bottom Left}: the distribution of local homophily counts on Photo. \tit{Bottom Right}: the distribution of local homophily ratios on Photo.}
    \label{fig: homo}    
\end{figure}

We compute global homophily ratios $\mca{H}_{global}$ on six real-world datasets, and explore the distributions of local homophily counts $\{\mca{H}_{local}^{\#}(u): u\in\mca{V}\}$ and ratios $\{\mca{H}_{local}(u): u\in\mca{V}\}$ on Amazon Photo. The statistics are illustrated in \Cref{fig: homo}. As can be observed: (a) the global homophily ratio varies considerably across datasets, with values ranging from 0.3195 (\tit{Flickr}) to 0.9314 (\tit{Physics}), indicating that it is ubiquitous that structural connections between neighbors do not coincide with their semantic relations; and (b) although local homophily is relatively more concentrated in smaller counts and greater ratios, the overall distribution is hard to estimate, let alone identify exact semantically positive neighbors in the context of unsupervised learning~\citep{sun2023progressive}. 
In a nutshell, overemphasizing neighborhood affinity in contrastive learning is risky since the model would be provided with noisy and even detrimental learning signals, which result in suboptimal performance. Motivated by the above findings, we consider \tit{soft neighborhood awareness} to realize single-view GCL.

% \begin{figure}
% \centering
% \begin{minipage}{.47\columnwidth}
%     \subfloat{
%       % \label{c} 
%       \includegraphics[width=.95\columnwidth]{figure/homo_ratio.pdf}}
% \end{minipage}
% \begin{minipage}{.5\columnwidth}
%     \subfloat{
%       % \label{a} 
%       \includegraphics[width=.98\columnwidth]{figure/number_per_node.pdf}}\\
%     \subfloat{
%       % \label{b} 
%       \includegraphics[width=.95\columnwidth]{figure/proportion_per_node.pdf}}
% \end{minipage}
% \caption{Homophily statistics from different perspectives. \tbf{Left}: homophily ratios on different datasets. \tbf{Upper Right}: the distribution of the number of same-label neighbors w.r.t. each node on Amazon Photo. \tbf{Lower Right}: the distribution of the proportion of same-label neighbors w.r.t. each node on Amazon Photo.}
% \label{fig: homo}
% \end{figure}

\section{Methodology} \label{sec: method}
% \begin{minipage}{0.58\textwidth}
%     \centering
%    \includegraphics[width=0.98\linewidth]{figure/framework.pdf}
%    \captionof{figure}{The framework of our proposed method. As for BaseEncoder layers, we utilize linear layers (which means that the adjacency matrix is not included in the input) and graph convolutional layers in transductive and inductive learning tasks, respectively.}
%    \label{fig: framework}
% \end{minipage}%
% \hfill % Fill the horizontal space between the minipages to push them apart

\begin{figure*}
   \centering   \includegraphics[width=0.98\linewidth]{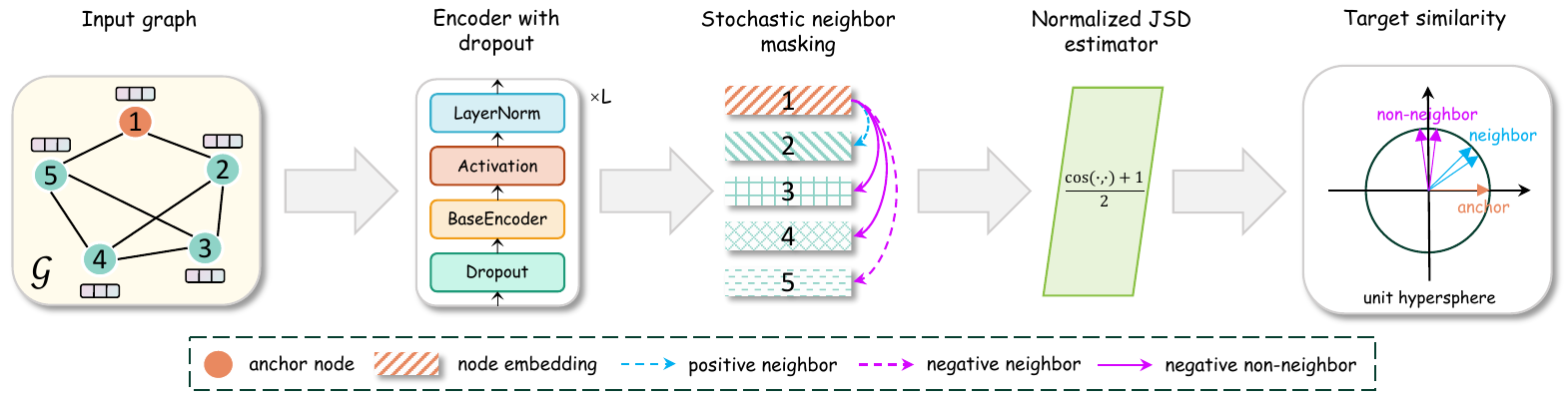}
   \caption{The framework of our proposed \ttt{SIGNA}. The contrast is conducted within a single graph view, and thus \ttt{SIGNA} has only one branch. The encoder with dropout implicitly provides more embedding combinations for robust contrast. The role of neighbors is variable, while non-neighbors are fixed as negative samples. The normalized JSD estimator facilitates better contrastive effect. The goal of \ttt{SIGNA} is to realize soft neighborhood awareness.}
   \label{fig: framework}
\end{figure*}
The overall framework of our proposed \ttt{SIGNA} is depicted in \Cref{fig: framework}. \ttt{SIGNA} employs a single-branch paradigm for single-view contrast, eliminating the need for augmented inputs, disparate encoders, or perturbed embeddings to generate multiple views. Simply pulling embeddings of neighbors and pushing away those of non-neighbors has shown to be less competitive~\citep{kipf2016GAE}.
To pursue soft neighborhood awareness, \ttt{SIGNA} is equiped with three main components, \tit{i.e.}, the encoder with dropout, the stochastic neighbor masking, and the normalized Jensen-Shannon divergence (Norm-JSD) estimator.

\tbf{Encoder with Dropout.} Dropout has been adopted as \tit{minimal data augmentation} in contrastive learning~\citep{gao2021simcse, xu2023simdcl}. To obtain positive pairs, they pass the same instance to the dropout-contained encoder twice so as to generate two \tit{related yet discrepant} embeddings. By contrast, we only pass the input graph to the encoder with dropouts \tit{ONCE}. In this way, the embedding pairs of neighbors are \tit{structurally related yet randomly noised}, making them competent for positive samples in single-view contrastive learning. Note that we do not rely on augmentations to generate contrastive pairs. The introduction of random noise just implicitly creates more diverse embedding combinations for contrast. Thus, the encoder is less likely to overfit the relationship between node pairs when conducting contrastive learning. 

Specifically, the encoder is composed of a stack of $L$ identical layers (in this paper, we fix $L=2$ to match the common depth of existing works). Each layer has a $\moprm{BaseEncoder}$ which can either be a linear layer (on transductive tasks) or a graph convolutional layer~\citep{kipf2016GCN} (on inductive tasks). 
% Accordingly, only node features are required as input data for a linear layer, while an extra adjacency matrix is in demand for a graph convolutional layer.
Before each $\moprm{BaseEncoder}$, we employ $\moprm{Dropout}$ (with rescaling)~\citep{srivastava2014dropout} to inject noise. The output of $\moprm{BaseEncoder}$ is further processed by a nonlinear $\moprm{Activation}$ function and a $\moprm{LayerNorm}$. Let $\mbf{H}^{(l-1)}$ be the input embeddings of $l$-th encoder layer (particularly, $\mbf{H}^{(0)}=\mbf{X}$), the forward encoding process within an encoder layer can be formally described as:
\begin{equation}
    \mbf{H}^{(l)}_\mrm{d} = \moprm{Dropout}(\mbf{H}^{(l-1)};p),
\end{equation}
\begin{equation}
    \mbf{H}^{(l)}_\mrm{e} = \moprm{Linear}(\mbf{H}^{(l)}_\mrm{d}) \ \ \text{or} \ \
    \mbf{H}^{(l)}_\mrm{e} = \moprm{GConv}(\mbf{H}^{(l)}_\mrm{d}, \mbf{A}),
\end{equation}
\begin{equation}
    \mbf{H}^{(l)}_\mrm{a}=\moprm{Activation}(\mbf{H}^{(l)}_\mrm{e}),
\end{equation}
\begin{equation}
    \mbf{H}^{(l)} = \mbf{H}^{(l)}_\mrm{n}=\moprm{LayerNorm}(\mbf{H}^{(l)}_\mrm{a}),
\end{equation}
where $p$ denotes the dropout rate, $\moprm{Linear}(\mbf{H}^{(l)}_\mrm{d})=\mbf{H}^{(l)}_\mrm{d}\mbf{W}^{(l)}$ denotes a linear layer, and $\moprm{GConv}(\mbf{H}^{(l)}_\mrm{d}, \mbf{A})=\mbf{\hat{D}}^{-\frac{1}{2}}\mbf{\hat{A}}\mbf{\hat{D}}^{-\frac{1}{2}}\mbf{H}^{(l)}_\mrm{d}\mbf{W}^{(l)}$ denotes a graph convolutional layer with $\mbf{\hat{A}}=\mbf{A}+\mbf{I}_N$ and $\hat{D}_{ii} = \sum_{j=1}^{N} {\hat{A}}_{ij}$. We denote the parameter set of the to-be-trained encoder by $\theta=\{\mbf{W}^{(l)}\}_{ l=1,\cdots,L}$. 
The output of the last encoder layer is our learned representations, \tit{i.e.}, $\mbf{H}=\mbf{H}^{(L)}$.

\tbf{Stochastic Neighbor Masking.} According to previous homophily analyses, neighbors actually have a non-negligible probability of owning different labels. In light of such uncertainty, we seek to mask a fraction of neighbors and consider the remaining neighbors as positive samples. To do that, we independently draw a masking indicator $m_v$ for a neighboring node $v$ of $u$ from a Bernoulli distribution with probability $\alpha\in [0,1]$, \tit{i.e.}, $m_v\sim\mca{B}(\alpha), \forall v\in \mca{N}_u$. Specifically, the remaining neighbor set at each epoch is:
\begin{equation}
    \mca{N}'_u=\{v\in\mca{N}_u\mid m_v=0\}.
\end{equation}
Just in case all neighbors are masked, we add the anchor node itself to its positive set for implementation convenience (no extra training signal is provided). As for negative set, it consists of all non-neighbors as well as those masked neighbors. Then, the positive and negative sets are:
\begin{equation}
    \mca{P}_u=\mca{N}'_u\cup\{u\}, \;  \mca{Q}_u=\mca{V}\setminus \mca{P}_u.
\label{eq: sample}
\end{equation}

\begin{restatable}[Probabilistic Neighborhood Contrastive Learning]{theorem}{PNCL} \label{the: pncl}
Let $S_{uv}$ be the target similarity between embeddings of the anchor node $u$ and any other node $v\neq u$ within the graph, and assume that $S_{uv}=\delta$ if $v\in \mca{P}_u$ otherwise $S_{uv}=\lambda$ ($v\in \mca{Q}_u$), where $\delta,\lambda$ are determined by the objective function. Then, we have: (a) $\mbb{E}_{v\in\mca{N}_u}(S_{uv})=\delta(1-\alpha)+\lambda\alpha$; (b) $\mbb{E}_{v\notin\mca{N}_u}(S_{uv})=\lambda$.
\end{restatable}
\begin{proof}
% \vspace{-0.2cm}     
    % According to \Cref{eq: loss}, $S_{uv}=1$ if $v\in \mca{P}_u$ otherwise $S_{uv}=0$ ($v\in \mca{Q}_u$). 
    (a) Since neighbors are randomly masked by a probability of $\alpha$, we have $p(v\in\mca{P}_u|v\in\mca{N}_u)=1-\alpha$ and $p(v\in\mca{Q}_u|v\in\mca{N}_u)=\alpha$. Thus, $\mbb{E}_{v\in\mca{N}_u}(S_{uv})=\delta(1-\alpha)+\lambda\alpha$. (b) Since $p(v\in\mca{Q}_u|v\notin\mca{N}_u)=1$, $\mbb{E}_{v\notin\mca{N}_u}(S_{uv})=\lambda$ holds.
    % \vspace{-0.2cm}
\end{proof}

\tit{Remark.} With stochastic masking, neighbors are flipped back and forth as positive and negative samples along the training process, while non-neighbors consistently act as negative samples. As a result, the expectation of target similarity scores among neighbors lies between $\delta$ and $\lambda$. 

\tbf{Normalized JSD Estimator.} According to the empirical evidence provided by \citep{hjelm2018DIM}, InfoNCE (Noise Contrastive Estimation~\cite{oord2018CPC}) and DV (Donsker-Varadhan representation of the KL-divergence~\cite{donsker1983DV}) require a large number of negative samples to be competitive, while JSD (Jensen-Shannon Divergence estimator~\cite{nowozin2016fGan}) is less sensitive to the number of negative samples. Since we sample positive samples from neighbors, the number of negative samples is inherently reduced. Therefore, we are committed to optimizing an objective in the form of JSD estimator:
\begin{equation}
\begin{split}
    \mca{J}_\mrm{JSD}(u)=&\mbb{E}_{v^+\sim \mca{P}_u}[\log \mca{D}_\phi(u,v^+)] \\
    &+\mbb{E}_{v^-\sim \mca{Q}_u}[\log (1-\mca{D}_\phi(u,v^-))],
\end{split}    
\label{eq: jsd}
\end{equation}
where $\mca{D}_\phi: \mbb{R}^d\times\mbb{R}^d\rightarrow \mbb{R}$ is a discriminator function modeled by a neural network with parameters $\phi$ to measure the similarity between two instances and scale it into the range of [0,1]. Typically, the discriminator of JSD estimator is implemented using the inner product plus sigmoid function~\citep{nowozin2016fGan, hjelm2018DIM, peng2020GMI}. Yet, it has been empirically verified that $\ell_2$ normalization plays an important role in contrastive learning~\citep{chen2020SIMCLR}, which projects embeddings to the unit hypersphere before computing similarity~\citep{wang2020understandingAU}. Hence, we introduce a normalized discriminator for the JSD estimator:
\begin{equation}
    \mca{D}_\phi^\mrm{norm}(u,v)=\frac{1}{2}\left(\frac{\mbf{z}_u^\top\mbf{z}_v}{{\|\mbf{z}_u \|}_2{\|\mbf{z}_v\|}_2}+1\right)=\frac{\moprm{cos}(\mbf{z}_u,\mbf{z}_v)+1}{2},
\label{eq: discriminator}
\end{equation}
where $\mbf{z}=g_\phi(\mbf{h})=\mbf{W}_g^{(2)}\sigma(\mbf{W}_g^{(1)}\mbf{h})$ is a MLP projector parameterized by $\phi=\{\mbf{W}_g^{(1)},\mbf{W}_g^{(2)}\}$. With $\ell_2$ normalization, the similarity metric between embedding vectors is calculated by cosine similarity instead of inner product. And we adopt a simple linear scaler to restrict the range of similarity scores. The instantiated estimator with the normalized discriminator is called \tit{Norm-JSD}. In \Cref{tab: Norm-JSD}, we compare Norm-JSd against two typical estimators, JSD and InfoNCE, which implies that Norm-JSD succeeds in combining the advantages of both JSD and InfoNCE while remaining concise. The superiority of Norm-JSD over JSD and InfoNCE is also empirically demonstrated in our ablation studies.

\begin{table}[t]
\centering
    {\small
    \begin{tabular}{lccc}
        \toprule
        Perspective &  InfoNCE & JSD & Norm-JSD  \\        
        \midrule
         Robust to $\#$neg.  & no & yes & yes\\ 
         \midrule
        {$\ell_2$ normalization}  & {yes} & no & yes \\
        \midrule
        Scaler & temperature & sigmoid & linear \\
        % \midrule
        % Linear scaler & \XSolidBrush & \XSolidBrush & \CheckmarkBold\\
        \bottomrule
    \end{tabular}}
    \caption{Comparing Norm-JSD against JSD and InfoNCE.}
    \label{tab: Norm-JSD}
\end{table}

Combining \Cref{eq: sample}, \Cref{eq: jsd}, and \Cref{eq: discriminator} together, we arrive at the loss function of \ttt{SIGNA}:
\begin{equation}
\label{eq: loss}
\begin{split}        
    {\ell}(u)= & -\frac{1}{|\mca{P}_u|}\sum\nolimits_{v^+\in\mca{P}_u}\log\left(\mca{D}_\phi^\mrm{norm}\left(u,{v^+}\right)\right) \\
    & -\frac{1}{|\mca{Q}_u|}\sum\nolimits_{v^-\in\mca{Q}_u}\log\left(1-\mca{D}_\phi^\mrm{norm}\left(u,{v^-}\right)\right),
\end{split}
\end{equation}
\begin{equation}
    \mca{L}_\mrm{SIGNA}=\frac{1}{|\mca{V}|}\sum\nolimits_{u\in\mca{V}}{\ell}(u).
\end{equation}

Note that the target similarity scores for positive and negative samples are $\mca{D}_\phi^\mrm{norm}\left(u,{v^+}\right)=1$ and $\mca{D}_\phi^\mrm{norm}\left(u,{v^-}\right)=0$, respectively. Namely, $\delta=1$ and $\lambda=0$. Recalling \Cref{the: pncl}, we have the following corollary:

\begin{restatable}[]{corollary}{corol} \label{corol}
With Norm-JSD, the expectation of target similarity for neighbors equals $1-\alpha$ and the expectation of target similarity for non-neighbors equals $0$, \tit{i.e.}, $\mbb{E}_{v\in\mca{N}_u}(S_{uv})=1-\alpha \in [0,1]$ and $\mbb{E}_{v\notin\mca{N}_u}(S_{uv})=0$.
\end{restatable}

\tit{Remark.} The above corollary indicates that, with soft neighborhood awareness, neighbors are expected to maintain a \tit{moderate} level of local tolerance with anchors, which leaves some leeway for future tuning in downstream tasks. By contrast, a large number of non-neighbors will be separated as far as possible. In other words, \ttt{SIGNA} is allowed to learn a desirable globally-uniform yet locally-tolerant embedding space~\citep{wang2020understandingAU, wang2021understandingUT}. Such an effect is observed in our experimental analysis.

% \begin{table}
%     \centering
%      \small
%     \begin{tabular}{lcc}
%         \toprule
%        Objectives & $l_2$ Norm  & Robust to $\#$negatives  \\
%         \midrule
%         InfoNCE & \CheckmarkBold & \XSolidBrush \\
%         \midrule
%          JSD  & \XSolidBrush & \CheckmarkBold \\
%         \midrule
%         JSD-LSC & \CheckmarkBold & \CheckmarkBold \\  
%         \bottomrule
%     \end{tabular}
%     \caption{Comparing JSD-LSC against typical JSD and InfoNCE.}
%     \label{tab: JSD_LSC}
% \end{table}

\section{Experiments} \label{sec: experiment}
% We conduct experiments to evaluate the proposed method by answering the following questions:
% \begin{itemize}
%     \item \tbf{RQ1}:
% \end{itemize}

\subsection{Experimental Setup}
\noindent\tbf{Datasets.} 
% Following previous works in self-supervised representation learning~\citep{velickovic2019DGI, jiao2020subg-con, thakoor2021BGRL, lee2022AFGRL}, 
We comprehensively evaluate \ttt{SIGNA} on three kinds of node-level tasks across 7 datasets with various scales and properties~\citep{velickovic2019DGI, jiao2020subg-con, thakoor2021BGRL, lee2022AFGRL}. \tit{Wiki CS}, \tit{Amazon Photo}, \tit{Amazon Computers}, \tit{Coauthor CS}, and \tit{Coauthor Physics} are used for transductive node classification and node clustering tasks. Two larger-scale datasets, \tit{Flickr} and \tit{PPI}, are used for inductive node classification on a single graph and multiple graphs, respectively. Statistics of these datasets are presented in the Appendix.

\noindent\tbf{Baselines.} We primarily compare \ttt{SIGNA} against representative and state-of-the-art unsupervised methods for node representation learning, including GAE~\citep{kipf2016GAE}, DGI~\citep{velickovic2019DGI}, GMI~\citep{peng2020GMI}, MVGRL~\citep{hassani2020MVGRL}, GRACE~\citep{zhu2020GRACE}, Subg-Con~\citep{jiao2020subg-con}, S$^2$GRL~\citep{peng2020s2grl}, CCA-SSG~\citep{zhang2021CCA-SSG}, BGRL~\citep{thakoor2021BGRL},  GraphCL-NS~\citep{hafidi2022graphcl-ns}, SUGRL~\citep{mo2022SUGRL}, and AFGRL~\citep{lee2022AFGRL}.

\noindent\tbf{Evaluation Protocols.} We first train all models in a fully unsupervised manner, and then the trained encoder is frozen and used for testing in downstream tasks (see Appendix). 
% For node classification tasks, we follow the linear evaluation scheme~\cite{velickovic2019DGI, zhu2020GRACE} to train and test a simple logistic regression classifier. We train the classifier for 20 runs with different data splits and report the micro-averaged F1-score. For node clustering tasks, we perform clustering on the learned representations using the K-means algorithm and report Normalized Mutual Information (NMI) and Homogeneity following~\citep{lee2022AFGRL}. % Unless specifically stated, the metrics of baselines are taken directly from their original papers or retrained with the open-source code. 

\noindent\tbf{Implementation Details.} Details about encoder implementation, hyperparameter selection, and computing infrastructure are provided in appendix due to the space limitation. 
% On all tasks, our encoder consists of two encoding blocks ($L=2$). For transductive node classification and node clustering, we use the simple linear layer as the BaseEncoder, while for inductive node classification, the graph convolutionial layer is adopted for better inference effect. All hyperparameters are selected via grid search and are specified in Appendix. 

\begin{table*}[!t]
\centering
{\small
\begin{tabular}{lcccccc}
\toprule
Method & Training Data & Wiki CS  & Amazon Photo & Amazon Computers & Coauthor CS & Coauthor Physics \\
\midrule
GCN & $\mathbf{X,A,Y}$ & 77.19$\pm$0.12 & 92.42$\pm$0.22 & 86.51$\pm$0.54 & {93.03$\pm$0.31} & {95.65$\pm$0.16}  \\
GAT & $\mathbf{X,A,Y}$ & {77.65$\pm$0.11} & {92.56$\pm$0.35} & {86.93$\pm$0.29} & 92.31$\pm$0.24 & 95.47$\pm$0.15 \\
\midrule
% Raw feats. & $\mathbf{X}$  & 71.98$\pm$0.00 & 78.53$\pm$0.00 & 73.81$\pm$0.00 & 90.37$\pm$0.00 & 93.58$\pm$0.00 \\
% node2vec & $\mathbf{A}$ & 71.79$\pm$0.05 & 89.67$\pm$0.12 & 84.39$\pm$0.08 &85.08$\pm$0.03 & 91.19$\pm$0.04  \\
% DeepWalk & $\mathbf{A}$ & 74.35$\pm$0.06 & 89.44$\pm$0.11 & 85.68$\pm$0.06 & 84.61$\pm$0.22 & 91.77$\pm$0.15 \\
% DW +feats. & $\mathbf{X,A}$ & 77.21$\pm$0.03 & 90.05$\pm$0.08 & 86.28$\pm$0.07 & 87.70$\pm$0.04 & 94.90$\pm$0.09 \\
% \midrule
GAE & $\mathbf{X,A}$ & 75.25$\pm$0.28 & 91.62$\pm$0.13 & 85.27$\pm$0.19 & 90.01$\pm$0.17 & 94.92$\pm$0.07 \\
VGAE & $\mathbf{X,A}$  & 75.63$\pm$0.19 & 92.20$\pm$0.11 & 86.37$\pm$0.21 & 92.11$\pm$0.09 & 94.52$\pm$0.00\\
DGI & $\mathbf{X,A}$ & 75.35$\pm$0.14 & 91.61$\pm$0.22 & 83.95$\pm$0.47 & 92.15$\pm$0.63 & 94.51$\pm$0.52  \\
GMI & $\mathbf{X,A}$ & 74.85$\pm$0.08 & 90.68$\pm$0.17 & 82.21$\pm$0.31 & OOM & OOM  \\
MVGRL & $\mathbf{X,A}$ & 77.52$\pm$0.08 & 92.08$\pm$0.01 & 87.52$\pm$0.21 & 92.18$\pm$0.15 & 95.33$\pm$0.03 \\
GRACE & $\mathbf{X,A}$ & 78.19$\pm$0.41 & 92.24$\pm$0.45 & 86.35$\pm$0.44 & 92.93$\pm$0.22 & 95.26$\pm$0.10 \\
CCA-SSG & $\mathbf{X,A}$ & 78.64$\pm$0.72 & 93.14$\pm$0.14 & 88.74$\pm$0.28 & 92.91$\pm$0.20 & 95.38$\pm$0.06 \\
SUGRL & $\mathbf{X,A}$ & 79.12$\pm$0.67  & 93.07$\pm$0.15 & 88.93$\pm$0.21 & 92.83$\pm$0.23 & 95.38$\pm$0.11 \\
BGRL & $\mathbf{X,A}$ & \udl{79.36$\pm$0.53}  & 92.87$\pm$0.27 & 89.68$\pm$0.31 & 93.21$\pm$0.18 & 95.56$\pm$0.12 \\
AFGRL & $\mathbf{X,A}$ & 77.62$\pm$0.74 & \udl{93.22$\pm$0.28} & \udl{89.88$\pm$0.33} & \udl{93.27$\pm$0.17} & \udl{95.69$\pm$0.08}  \\
% Local-GCL & $\mathbf{X,A}$ & - & \udl{93.25$\pm$0.40} & 88.81$\pm$0.37 & \udl{94.90$\pm$0.19} & \udl{96.33$\pm$0.13} \\
\ttt{SIGNA} & $\mathbf{X,A}$ & \tbf{80.91$\pm$0.46} & \textbf{95.32$\pm$0.19} & \textbf{90.46$\pm$0.25} & \textbf{94.98$\pm$0.20} & \textbf{96.35$\pm$0.09}  \\
\bottomrule
\end{tabular}}
\caption{Performance on transductive node classification tasks.
% in terms of averaged accuracy with standard deviation. 
% Training data indicates the available input data during training. 
% OOM denotes out of memory on a GPU with 32GB of memory. 
The best and second-best performances among unsupervised methods are highlighted in bold and underlined, respectively.}
\label{tab: transductive_results}
\end{table*}

\subsection{Main Results and Analysis}
\tbf{Transductive Node Classification.} The empirical performance of various methods across five datasets are summarized in \Cref{tab: transductive_results}. As demonstrated by the results, \ttt{SIGNA} consistently and substantially outperforms baseline methods across all five benchmark datasets. Specifically, \ttt{SIGNA} improves the accuracy by absolute margins of $1.55\%$ (\tit{Wiki CS}), 2.04\% (\tit{Amazon Photo}), $0.58\%$ (\tit{Amazon} \tit{Computers}), $1.71\%$ (\tit{Coauthor CS}), and $0.66\%$ (\tit{Coauthor Physics}) over those competitive runner-ups. Besides, a notable finding is that the performances of those competitors are less stable across different benchmarks. For example, AFGRL~\citep{lee2022AFGRL} outperforms BGRL~\citep{thakoor2021BGRL} on four out of five datasets while being surpassed by BGRL~\citep{thakoor2021BGRL} with a margin of $1.74\%$ on \tit{Wiki CS}. By contrast, our method achieves more robust leading performances, indicating the efficacy and universality of \ttt{SIGNA}. Furthermore, unlike existing methods that rely on graph convolutional layers to generate representations, simple linear layers are used
as our BaseEncoders, which discard the complicated massage aggregation process on numerous edges. To show the benefit, we investigate the time cost of the inference phase with GCN and MLP being basic encoders (all other settings are kept consistent), which are reported in \Cref{tab: inference time}. As illustrated, our MLP-based encoder is about $109\times$ to $331\times$ faster than the GCN-based encoder with the same settings in terms of the inference time. In particular, the gap widens rapidly as the size of the edge set increases, since the inference of MLP-based encoders is independent of edge-oriented aggregation.  

\begin{table}[!t]
    \centering
    {\small
    \begin{tabular}{lccccc}
        \toprule
        {Encoder} & Wiki CS & Photo & Comp. & CS & Phys. \\
        \midrule
        GCN & 37.67 & 30.62 & 60.75 & 62.85 & 145.70 \\
        MLP & \tbf{0.25} & \tbf{0.28} & \tbf{0.31} & \tbf{0.35} & \tbf{0.44}\\
        \midrule\midrule
        Ratio & $151\times$ & $109\times$ & $196\times$ & $180\times$ & $331\times$\\ 
        \bottomrule
    \end{tabular}  }
    \caption{Inference time (millisecond) on transductive node classification tasks with different BaseEncoders.}
    \label{tab: inference time} 
\end{table}
\begin{table}[!t]   
\centering
    {\small
    \begin{tabular}{lcc}
        \toprule
        Method & Training Data &  F1-score \\
        \midrule
        FastGCN & $\mathbf{X,A,Y}$ & 48.1$\pm$0.5\\
        GCN & $\mathbf{X,A,Y}$ & 48.7$\pm$0.3 \\
        GraphSAGE & $\mathbf{X,A,Y}$ & 50.1$\pm$1.3 \\
        \midrule
        % Raw features & $\mathbf{X}$ & 20.3$\pm$0.2 \\
        % DeepWalk & $\mathbf{A}$ & 27.9 \\
        % \midrule
        Unsup-GraphSAGE & $\mathbf{X,A}$ & 36.5$\pm$1.0\\
        DGI & $\mathbf{X,A}$ & 42.9$\pm$0.1\\
        GMI & $\mathbf{X,A}$ & 44.5$\pm$0.2 \\
        Subg-Con & $\mathbf{X,A}$ & \udl{48.8$\pm$0.1}\\     
        \ttt{SIGNA} & $\mathbf{X,A}$ & \textbf{51.93$\pm$0.04} \\
        \bottomrule
    \end{tabular}   } 
    \caption{Performance on single-graph inductive node classification task (Flickr) in terms of micro-averaged F1-score.}
    \label{tab: flickr}
\end{table}   
\begin{table}[!t]
    \centering
    {\small   
    \begin{tabular}{lcc}
        \toprule
        Method & Training Data &  F1-score \\
        \midrule
        GaAN-mean & $\mathbf{X,A,Y}$ & 96.9$\pm$0.2\\
        GAT & $\mathbf{X,A,Y}$ & 97.3$\pm$0.2\\
        FastGCN & $\mathbf{X,A,Y}$ & 63.7$\pm$0.6\\
        \midrule
        % Raw features & $\mathbf{X}$ & 42.2 \\   
        % DeepWalk & $\mathbf{A}$ & 52.9 \\
        % \midrule
        Unsup-GraphSAGE & $\mathbf{X,A}$ & 46.5\\
        Random-Init & $\mathbf{X,A}$ & 62.6$\pm$0.2\\
        DGI & $\mathbf{X,A}$ & 63.8$\pm$0.2\\
        GMI & $\mathbf{X,A}$ & 64.6$\pm$0.0 \\
        S$^2$GRL & $\mathbf{X,A}$ & 66.0$\pm$0.0 \\
        GRACE & $\mathbf{X,A}$ & 66.2$\pm$0.1\\
        Subg-Con & $\mathbf{X,A}$ & {66.9$\pm$0.2}\\  
        GraphCL-NS & $\mathbf{X,A}$  & 65.9$\pm$0.0\\
        BGRL-GAT-Encoder & $\mathbf{X,A}$  & \udl{70.49$\pm$0.05}\\
        \ttt{SIGNA} & $\mathbf{X,A}$ & \textbf{92.25$\pm$0.03} \\
        \bottomrule
    \end{tabular}}
     \caption{Performance on multi-graph inductive node classification task (PPI) in terms of micro-averaged F1-score.}
    \label{tab: ppi}
\end{table}

\noindent\tbf{Inductive Node Classification.} The empirical performances on a single graph (\tit{Flickr}) and multiple graphs (\tit{PPI}) are reported in \Cref{tab: flickr} and \Cref{tab: ppi}, respectively. According to the results in terms of micro-averaged F1-score, \ttt{SIGNA} surpasses the previous best method by $3.13\%$ (\tit{Flickr}) and $21.74\%$ (\tit{PPI}), showcasing its remarkable superiority in inductive learning tasks. It is noteworthy that \ttt{SIGNA} even outperforms the best supervised baseline GraphSAGE~\citep{hamilton2017graphsage} by $1.83\%$ on \tit{Flickr}. On the \tit{PPI} dataset, while the performance of supervised baselines remains unsurpassed, our method significantly reduces the disparity between unsupervised approaches and those supervised benchmarks. 

\begin{table*}
    \centering
    {\small
    \begin{tabular}{lccccccccccc}
        \toprule
        \multirow{2}{*}{Method} & \multicolumn{2}{c}{Wiki CS} & \multicolumn{2}{c}{Am. Photo} & \multicolumn{2}{c}{Am. Computers} & \multicolumn{2}{c}{Co. CS} & \multicolumn{2}{c}{Co. Physics} \\
        \cmidrule(lr){2-3} \cmidrule(lr){4-5} \cmidrule(lr){6-7} \cmidrule(lr){8-9} \cmidrule(lr){10-11} 
         & NMI & Homo. & NMI & Homo. & NMI & Homo. & NMI & Homo. & NMI & Homo. \\
        \midrule
        Raw features & 0.2633 & 0.2738 & 0.3273 & 0.3376& 0.2389 & 0.2617 & 0.7103 & 0.7446 & 0.5207 & 0.5576\\
        GRACE & \udl{0.4282} & \udl{0.4423} & 0.6513 & 0.6657 & 0.4793 & 0.5222 & 0.7562 & 0.7909 & 0.5128 & 0.5546 \\
        GCA & 0.3373 & 0.3525 & 0.6443 & 0.6575 & 0.5278 & 0.5816 & 0.7620 & 0.7965 & 0.5202 & 0.5654 \\
        BGRL & 0.3969 & 0.4156 & \udl{0.6841} & \udl{0.7004} & 0.5364 & 0.5869 & 0.7732 & 0.8041 & 0.5568 & 0.6018 \\
        AFGRL & 0.4132 & 0.4307 & 0.6563 & 0.6743 & \udl{0.5520} & \udl{0.6040} & \udl{0.7859} & \udl{0.8161} & \udl{0.5782} & \udl{0.6174} \\
        \ttt{SIGNA}  & \tbf{0.4593} & \tbf{0.4763} & \tbf{0.7635} & \tbf{0.7823} & \tbf{0.5608} & \tbf{0.6057} & \tbf{0.8047} & \tbf{0.8408} & \tbf{0.5907} & \tbf{0.6393} \\
        \bottomrule
    \end{tabular}}
    \caption{Performance on node clustering tasks in terms of NMI and Homogeneity.}
    \label{tab: clustering}
\end{table*}

\begin{table*}
    \centering
    {\small   
    \begin{tabular}{lccccc}
        \toprule   
        {Variant} &  Wiki CS & Am. Photo & Am. Computers & Co. CS & Co. Physics \\
        \midrule
        \ttt{SIGNA}  &\tbf{80.91$\pm$0.46} & \textbf{95.32$\pm$0.19} & \textbf{90.46$\pm$0.25} & \textbf{94.98$\pm$0.20} & \textbf{96.35$\pm$0.09} \\      
        \midrule
        \; $\times$Dropout  & 80.04$\pm$0.55 & 93.85$\pm$0.26 & 89.72$\pm$0.36 & 93.95$\pm$0.15 & 95.93$\pm$0.12 \\
        \; Dropout$\Rightarrow$NFM  & 75.21$\pm$0.45 & 91.16$\pm$0.53 & 87.43$\pm$0.36 & 93.92$\pm$0.20 & 95.95$\pm$0.09\\
        \midrule
        \; $\times$StochMask   & 80.58$\pm$0.65 & 94.87$\pm$0.28 & 90.09$\pm$0.34 & 93.74$\pm$0.15 & 96.22$\pm$0.08 \\
        \; StochMask$\Rightarrow$AllMask  & 68.14$\pm$0.69 & 80.59$\pm$0.99 & 76.64$\pm$0.60 & 89.22$\pm$0.25 & 94.30$\pm$0.14 \\   
        \midrule
                \; Norm-JSD$\Rightarrow$JSD  & 78.11$\pm$0.59 & 76.90$\pm$0.54 & 70.02$\pm$0.77 & 91.84$\pm$0.18 & 92.42$\pm$0.13 \\
        \; Norm-JSD$\Rightarrow$InfoNCE & 79.03$\pm$0.71 & 93.64$\pm$0.39 & 88.10$\pm$0.35 & 94.88$\pm$0.15 & 95.96$\pm$0.11 \\
        \midrule
        \; $\times$All & 77.92$\pm$0.48 & 74.61$\pm$1.22 & 69.16$\pm$0.83 & 86.74$\pm$0.34 & 50.57$\pm$0.10 \\  
        \bottomrule
    \end{tabular}}
\caption{Ablation study of \ttt{SIGNA}. $\times$: remove components; $\Rightarrow$: replace components with alternatives.}
    \label{tab: ablation}
\end{table*}

\noindent\tbf{Node Clustering.} The comparison of clustering performance against GRACE~\citep{zhu2020GRACE}, GCA~\citep{zhu2021GCA}, BGRL~\citep{thakoor2021BGRL}, and AFGRL~\citep{lee2022AFGRL} is shown in \Cref{tab: clustering}. The closer the value of both NMI and Homogeneity is to 1, the better the clustering effect. As revealed by \Cref{tab: clustering}, \ttt{SIGNA} exhibits dominating performance over all competitors across five datasets. In particular, \ttt{SIGNA} consistently outperforms AFGRL~\citep{lee2022AFGRL}, a clustering-oriented method, 
demonstrating the superior capability of \ttt{SIGNA} to learn underlying cluster structure in the unsupervised context.
% We attribute the success of \ttt{SIGNA} to the reduced reliance on the unverifiable neighbor infinity so as to strike a better balance between intra-class aggregation and inter-class separation, which will be further discussed via visualization.

\subsection{Why Does \ttt{SIGNA} Work}\label{subsec: why_work}

\begin{figure}
  \centering   \includegraphics[width=0.95\linewidth]{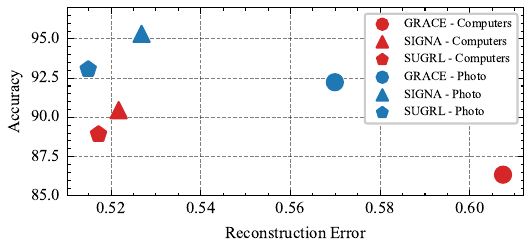}
   \caption{The soft neighborhood awareness of \ttt{SIGNA} yields better performance.}
   \label{fig: rec_acc}
\end{figure}

To facilitate direct understanding of what \ttt{SIGNA} has done, we first analyze the relationship between neighborhood awareness and model performance, which is illustrate in \Cref{fig: rec_acc}. Compared with representative neighborhood-aware (SUGRL) and non-neighborhood-aware (GRACE) GCL methods, \ttt{SIGNA} learns a moderate of structural information due to our soft neighborhood awarness strategies, thus yielding better generalization performances on downstream tasks. Besides, our histogram analysis reveals that pulling together neighbors is less rational than pushing away non-neighbors, and thus we should avoid overly pulling together neighbors, which exactly matches the spirit of \ttt{SIGNA}. Please refer to our appendix for details.

% \begin{figure}
%   \centering
%   % \vspace{-0.3cm}
%   % \begin{subfigure}{0.24\linewidth}
%   % \includegraphics[width=\linewidth]{figure/orig_tsne.pdf}
%   %   % \fbox{\rule{0pt}{2in} \rule{.9\linewidth}{0pt}}
%   %   \caption{Raw features}
%   %   \label{fig: tsne-a}
%   % \end{subfigure}
%   % \hfill
%   \begin{subfigure}{0.325\linewidth}
%     \includegraphics[width=\linewidth]{figure/grace_tsne.pdf}
%     \caption{GRACE}
%     \label{fig: tsne-b}
%   \end{subfigure}
%     \hfill
%   \begin{subfigure}{0.325\linewidth}
%     \includegraphics[width=\linewidth]{figure/sugrl_tsne.pdf}
%     \caption{SUGRL}
%     \label{fig: tsne-c}
%   \end{subfigure}
%     \hfill
%   \begin{subfigure}{0.325\linewidth}
%     \includegraphics[width=\linewidth]{figure/learned_tsne.pdf}
%     \caption{\ttt{SIGNA}}
%     \label{fig: tsne-d}
%   \end{subfigure}
%   \caption{Representation visualization via t-SNE (with the same random seed) on Amazon Photo. Colors indicate labels.}
%   \label{fig: tsne}
% \end{figure}

% \begin{figure}
% % \vspace{-0.3cm}
%   \centering
%   \begin{subfigure}{0.49\linewidth}
%     \includegraphics[width=\linewidth]{figure/dropout_rate.pdf}
%     % \caption{GRACE}
%     % \label{fig: tsne-b}
%   \end{subfigure}  
%     \hfill
%   \begin{subfigure}{0.49\linewidth}
%     \includegraphics[width=\linewidth]{figure/masking_rate.pdf}
%     % \caption{\ttt{SIGNA}}
%     % \label{fig: tsne-d}
%   \end{subfigure}
%   \caption{Sensitiveness of \ttt{SIGNA} to core hyperparameters. \tbf{Left}: the dropout rate ($p$); \tbf{Right}: the masking rate ($\alpha$).}
%   \label{fig: hyperpara}
% \end{figure}

\begin{figure}[!t]    
    % First group of three images
    \begin{minipage}{0.47\textwidth}        
        % Image 1
        \subfigure[GRACE]{\includegraphics[width=0.325\linewidth]{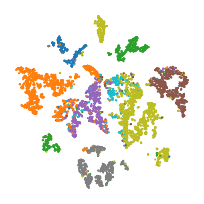}}
        \subfigure[SUGRL]{\includegraphics[width=0.325\linewidth]{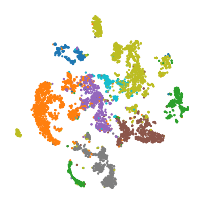}}
        \subfigure[\ttt{SIGNA}]{\includegraphics[width=0.325\linewidth]{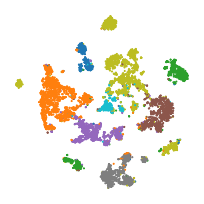}}
        \caption{Representation visualization via t-SNE.}
        \label{fig: tsne}
    \end{minipage}    
\end{figure}

In \Cref{fig: tsne}, we use t-SNE~\citep{van2008tsne} to visualize representations learned by \ttt{SIGNA}, as well as those learned by GRACE~\citep{zhu2020GRACE} and SUGRL~\citep{mo2022SUGRL} (representatives of augmentation-invariant and neighbourhood-aware methods). For fairness comparison, their random seeds are kept consistent. As can be seen, GRACE learns a uniformly-distributed embedding space, which impairs intra-class aggregation (e.g., purple samples). On the contrary, SUGRL improves intra-class similarities yet fails to separate different classes (e.g., purple samples and their surroundings). By contrast, \ttt{SIGNA} strikes a better balance between intra-class aggregation and inter-class separation (e.g., purple samples and their surroundings). Clearly, this ability contributes to its superiority in classification and clustering tasks.    

\subsection{Ablation Study} 
\label{sec: ablation}

In \Cref{tab: ablation}, we evaluate the contribution of different components of \ttt{SIGNA}. Overall, soft neighborhood awareness proves of great significance. Our three components are all reasonably designed, as removing any one of them results in a noticeable performance degradation, especially when all three are removed, the performance drops dramatically. Besides, our specific findings include: \tit{(i)} when we replace dropout with node feature masking (NFM), a widely-used input augmentation technique~\citep{zhu2020GRACE, zhu2021GCA, jin2021merit}, the performance decreases dramatically; \tit{(ii)} our Norm-JSD surpasses two other estimators by large margins, while the performance of JSD lags far behind that of InfoNCE, indicating the importance of our normalized discriminator; \tit{(iii)} 
masking all neighbors significantly reduces the performance due to the missing of structural information (recall that we use MLP as encoder for transductive tasks); Yet, overemphasizing neighborhood affinities without stochastic masking also yields an inferior performance. In short, our soft neighborhood awareness is more suitable for GCL.

% We consider the following variants. (a) \ttt{SIGNA-Neighbor}: removes all neighbors from the positive set of anchor nodes, which means that anchor nodes are required to be as dissimilar as possible from all other nodes (b) \ttt{SIGNA-Dropout}: removes the dropout layers within the encoder. (c) \ttt{SIGNA-LinScaCos}: replaces the linearly-scaled cosine similarity with the inner product scaled with sigmoid function, which is widely used in generative objectives~\citep{kipf2016GAE}. (d) \ttt{SIGNA-StocMask}: removes the stochastic masking scheme for neighbors, \tit{i.e.}, all neighbors are asked to be similar with anchor nodes. The results are reported in \Cref{tab: ablation}. Our key findings are: (1) On the one hand, completely discarding semantic information between neighbors (\ttt{SIGNA-Neighbor}) leads to a dramatic performance degradation. (2) On the other hand, alleviating over-reliance on structural information proves of great significance as the removal of any one of our key components results in a reduction in performance. In particular, the remarkable drop in the performance of \ttt{SIGNA-LinScaCos} illustrates the rationality of modelling structural affinity with inter-vector angles (cosine) instead of restricting both lengths and angles of representation vectors (inner product). Besides, the use of the Dropout technique (\ttt{SIGNA-Dropout}) accounts more than the stochastic masking of neighbors (\ttt{SIGNA-StocMask}) for the success of \ttt{SIGNA}.

\section{Conclusion}
In this work, we resort to soft neighborhood awareness for single-view GCL. To begin with, we explore homophily statistics on real-world datasets, which reveal that overconfidence in neighborhoods would be risky. Motivated by this, we propose a simple yet effective GCL framework (\ttt{SIGNA}) to pursue a moderate level of neighborhood affinities. 
Specifically, \ttt{SIGNA} considers neighboring nodes as potential positive samples but is equipped with three nuanced designs to relax restrictions on the retention of structural information. 
Experimental results demonstrate that \ttt{SIGNA} consistently outperforms existing GCL methods in a wide variety of node-level tasks. This work is expected to serve as a pioneer in exploring reasonable neighborhood awareness for contrastive learning of node representation.

% \begin{ack}
% Use unnumbered first level headings for the acknowledgments. All acknowledgments
% More information about this disclosure can be found at: \url{https://neurips.cc/Conferences/2024/PaperInformation/FundingDisclosure}.

% Do {\bf not} include this section in the anonymized submission, only in the final paper. You can use the \texttt{ack} environment provided in the style file to automatically hide this section in the anonymized submission.
% \end{ack}

\section{Acknowledgments}
The work of Ziyue Qiao was supported by National Natural Science Foundation of China under Grant No. 62406056. The work of Kai Wang was supported in part by National Natural Science Foundation of China under Grant No. U24A20270 and Grant No. 62373378.

% references
% \small
\bibliography{aaai25.bib}
% \bibliographystyle{plainnat}
% \bibliographystyle{plainnat}

%%%%%%%%%%%%%%%%%%%%%%%%%%%%%%%%%%%
%%%%%%       appendix        %%%%%%
%%%%%%%%%%%%%%%%%%%%%%%%%%%%%%%%%%%
\cleardoublepage
\onecolumn  % 切换到单列模式
\appendix   % 开始附录部分

\newpage
% \twocolumn
\appendix

\section*{APPENDIX}

\section{More Experimental Settings} \label{sup: exp}

Here, we provide more details about our experimental implementations, including dataset statistics, specified hyperparameters, and computing infrastructure.

% \subsection{Datasets} \label{sup: datasets}

\tbf{Datasets.} All datasets are available online from PyTorch Geometry (\url{https://pytorch-geometric.readthedocs.io/en/latest/modules/datasets.html}) with the MIT license. The statistics of datasets used in this work are summarized in \Cref{tab: statis}. Note that for \tit{Amazon Photo}, \tit{Amazon Computers}, \tit{Coauthor CS}, and \tit{Coauthor Physics}, we follow previous works~\citep{zhu2021GCA, thakoor2021BGRL} to randomly split the nodes into training/validation/testing
sets by the ratio of $10\%/10\%/80\%$ for linear evaluation, and report averaged results for 20 runs. 

For \tit{Wiki CS}, we use all 20 different splits shipped with the dataset and report averaged metrics for standard evaluation. 
For \tit{Flickr} and \tit{PPI}, we use the fixed official splits for training, validation, and testing. In particular, \tit{PPI} is a multi-graph dataset containing 24 graphs in total, which are split into training, validation, and testing set by 20/2/2 graphs.

\begin{table*}[h]
\centering
{\small
\begin{tabular}{lccccccc}
\toprule
Dataset & Domain & Task & \#Nodes & \#Edges & \#Features & \#Classes & Train/Val/Test\\
\midrule
Wiki CS & Knowledge base & TNC/NC &11,701 & 216,123 & 300 & 10 & 0.1/0.1/0.8\\
Ama. Photo & Social network & TNC/NC & 7,650 & 119,081 & 745 & 8 & 0.1/0.1/0.8\\
Ama. Comp. & Social network & TNC/NC & 13,752 & 245,861 & 767 & 10 & 0.1/0.1/0.8\\
Coa. CS & Citation network & TNC/NC & 18,333 & 81,894 & 6,805 & 15 & 0.1/0.1/0.8\\
Coa. Phys. & Citation network & TNC/NC & 34,493 & 247,962 & 8,415 & 5 & 0.1/0.1/0.8\\
\midrule
Flickr & Social network & INC & 89,250 & 899,756 & 500 & 7 & 44,625/22,312/22,313\\
\multirow{2}{*}{PPI} & \multirow{2}{*}{Protein network} & \multirow{2}{*}{INC}  & 56,944 & \multirow{2}{*}{806,174} & \multirow{2}{*}{50} & 121 & 44,906/6,514/5,524\\
 & & & (24 graphs) & & & (multi-label) & (20/2/2 graphs)\\
\bottomrule
\end{tabular}}
\caption{Summary of datasets used in experiments. TNC: transductive node classification. INC: inductive node classfication. NC: node clustering. The ratio-specified splits denote randomly splitting datasets by fixed ratios, and the number-specified splits denote standard and fixed splits of datasets for training, validation, and testing.}
\label{tab: statis}
\end{table*}

% \subsection{Evaluation protocol} 
\tbf{Evaluation Protocol.} We first train all models in a fully unsupervised manner, and then the trained encoder is frozen and used for testing in downstream tasks. For node classification tasks, we follow the linear evaluation scheme~\cite{velickovic2019DGI, zhu2020GRACE} to train and test a simple logistic regression classifier. We train the classifier for 20 runs with different data splits and report the micro-averaged F1-score. For node clustering tasks, we perform clustering on the learned representations using the K-means algorithm and report Normalized Mutual Information (NMI) and Homogeneity following~\citep{lee2022AFGRL}. % 

% \subsection{Encoder implementation}
\tbf{Encoder Implementation.} On all tasks, our encoder consists of two encoding blocks ($L=2$). For transductive node classification and node clustering, we use the simple linear layer as the BaseEncoder, while for inductive node classification, the graph convolutionial layer is adopted for better inference effect.  

% \subsection{Specified hyperparameters} \label{sup: hyperpara}
\tbf{Specified Hyperparameters.} All hyperparameters are selected via grid search. To guarantee the reproducibility of our performance, we specify hyperparamenters of \ttt{SIGNA} used in different datasets in \Cref{tab: hyperparameters}.

\begin{table*}[h]
% \vspace{-0.3cm}
% \vskip 0.15in
\centering
{\small
\begin{tabular}{l|cccccc|cc|c|ccc}
\toprule
Dataset & $p_{drop}$ & $\moprm{BEnc}$ & $\sigma_{enc}$ & $\moprm{LN}$ & $d_{enc}$ & $L$  & $d_{proj}$ & $\sigma_{proj}$ & $\alpha_{mask}$ & $lr$ & \#Epoch  & Opt.  \\
\midrule
Wiki CS & 0.4 & $\moprm{Linear}$ & PReLU & \XSolid & 1024 & 2 & 256 & ELU & 0.3 & 0.001 & 1500 & Adam \\
Photo & 0.4 & $\moprm{Linear}$ & PReLU & \Checkmark & 1024 & 2 & 256 & ELU & 0.4 & 0.001 & 1000 & Adam \\
Comp. & 0.3 & $\moprm{Linear}$ & PReLU & \Checkmark & 1024 & 2 & 256 & ELU & 0.8 & 0.001 & 3000 & Adam\\
CS & 0.7 & $\moprm{Linear}$ & PReLU & \Checkmark & 1024 & 2 & 512 & ELU & 0.95 & 0.0001 & 1500 & Adam \\
Phys. & 0.55 & $\moprm{Linear}$ & PReLU & \Checkmark & 1024 & 2 & 1024 & ELU & 0.8 & 0.0005 & 500 & Adam\\
\midrule
Flickr & 0.1  & $\moprm{GConv}$ & RReLU & \Checkmark & 1024 & 2 & 512 & ELU & 0.7 & 0.0005 & 50 & Adam\\
PPI & 0.4  & $\moprm{GConv}$ & ReLU & \Checkmark & 2048 & 2 & 1024 & ELU & 0.5 & 0.001 & 1500 & Adam\\
\bottomrule
\end{tabular}}
\caption{Specified hyperparameters on different datasets. $\moprm{BEnc}$: $\moprm{BaseEncoder}$. $p_\mrm{drop}$: probability of dropout an embedding dimension. $\sigma$: activation functions for encoder or projector. $d$: dimension of hidden embeddings in encoder or projector. $\alpha_\mrm{mask}$: probability of masking a neighbor. $lr$: learning rate. Opt.: optimizer}  
\label{tab: hyperparameters}
\end{table*}

% \subsection{Computing infrastructure} \label{sup: infra}
\tbf{Computing Infrastructure.} All experiments are conducted on a compute node with thirty-two Intel Xeon Gold 6242 CPUs (2.80GHz) and one Tesla
V100-SXM2 GPUs (32GB of memory).

\section{More Experimental Results}

% \subsection{Histogram analysis} 
Due to the space limitation, we report additional experimental results here, including training time comparison, histogram analysis and hyperparameter sensitiveness analysis.

\tbf{Training Time Comparison.} Here, we additionally report the training time of \ttt{SIGNA} with different encoders in transductive tasks. As illustrated in \Cref{tab: training time}, our MLP-based training times are also shorter than GCN-based training times, although the gaps are not as significant as the gaps in inference times.

\begin{table}[h]
    \centering
    {\small
    \begin{tabular}{lccccc}
        \toprule
        {Encoder} & Wiki CS & Photo & Comp. & CS & Phys. \\
        \midrule
        GCN & 166067 &	65625 &	420567 & 256917 & 312974 \\
        MLP & \tbf{91523} & \tbf{25865} & \tbf{237591} & \tbf{209314} & \tbf{274118}\\
        \midrule\midrule
        Ratio & $1.814\times$ & $2.537\times$ & $1.770\times$ & $1.227\times$ & $1.142\times$\\ 
        \bottomrule
    \end{tabular}  }
    \caption{Training time (millisecond) on transductive node classification tasks with different BaseEncoders.}
    \label{tab: training time} 
\end{table}

\begin{figure*}[h]
% \vspace{-0.5cm}
    % First group of three images
    \centering
    \begin{minipage}{0.495\textwidth}  
    \centering
        % Image 1
        \subfigure[Original]{\includegraphics[width=0.325\linewidth]{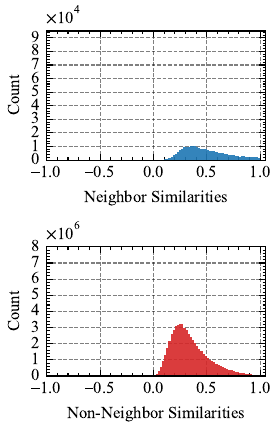}}
        \subfigure[SUGRL]{\includegraphics[width=0.325\linewidth]{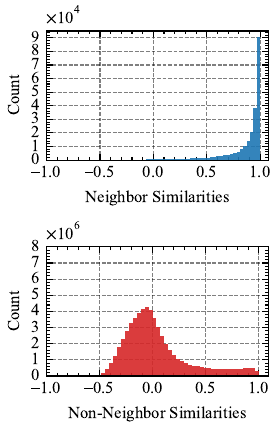}}
        \subfigure[\ttt{SIGNA}]{\includegraphics[width=0.325\linewidth]{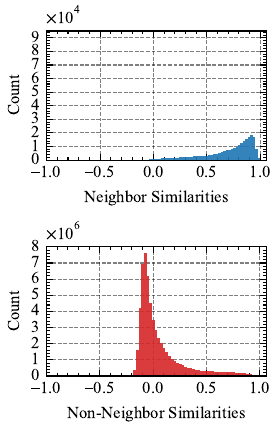}}
        \caption{Histograms of neighbor and non-neighbor similarities computed from features/representations on Photo.}  
        \label{fig: neigh}
    \end{minipage}    
    % Second group of three images
    \begin{minipage}{0.495\textwidth}      \centering
        % Image 1
        \subfigure[Original]{\includegraphics[width=0.325\linewidth]{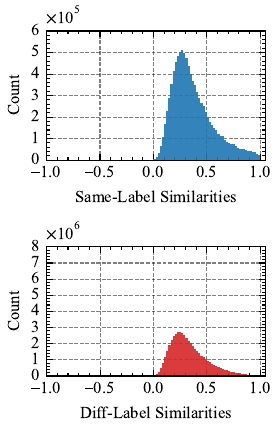}}
        \subfigure[SUGRL]{\includegraphics[width=0.325\linewidth]{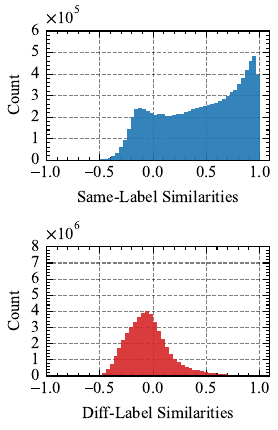}}
        \subfigure[\ttt{SIGNA}]{\includegraphics[width=0.325\linewidth]{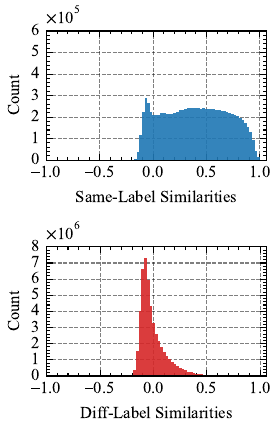}}
        \caption{Histograms of same-label and different-label similarities computed from features/representations on Photo.}
        \label{fig: label}
    \end{minipage}
\end{figure*}

\tbf{Histogram Analysis.} We plot histograms of pair-wise cosine similarities computed from learned representations w.r.t. neighbors vs. non-neighbors (\Cref{fig: neigh}) and same-label pairs vs. different-label pairs (\Cref{fig: label}). For reference, we also plot histograms of raw features and the embeddings learned by SUGRL. As can be seen in \Cref{fig: neigh}, \ttt{SIGNA} learns a relatively \tit{moderate} level of neighbor similarities than SUGRL, while its non-neighbor similarities are more significantly reduced than that of SUGRL. In \Cref{fig: label}, \ttt{SIGNA} succeeds in squeezing same-label similarities into a narrower interval ($[-0.15,1.0]$) than SUGRL ($[-0.5, 1.0]$), and its different-label similarities concentrate more around zero score. An interesting finding is that the overall distribution of neighbor similarities shows no resemblance to that of same-label similarity, while non-neighbor similarities share a highly consistent distribution with different-label similarities, implying a large number of overlaps between non-neighbors and different-label pairs. To this premise, a rational strategy is to focus more on separating non-neighbors rather than overly gathering neighbors, which exactly matches the spirit of our \ttt{SIGNA} and thus accounts for its efficacy.   

\begin{figure*}[h]
\centering
    \begin{minipage}{0.65\textwidth}        
        % Image 1
        \subfigure[$p$]{\includegraphics[width=0.49\linewidth]{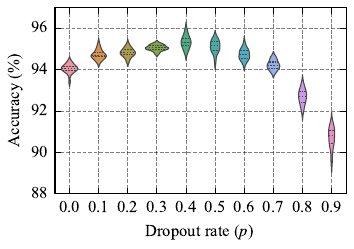}}
        \subfigure[$\alpha$]{\includegraphics[width=0.49\linewidth]{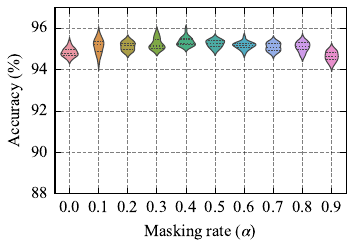}}  
        \caption{Hyperparameter sensitiveness.}   
    \label{fig: hyperpara}
    \end{minipage}
\end{figure*}
% \subsection{Hyperparameter sensitiveness analysis}
\tbf{Hyperparameter Sensitiveness Analysis.} In \Cref{fig: hyperpara}, we study the sensitiveness of \ttt{SIGNA} to its two core parameters, \tit{i.e.}, the embedding dropout rate ($p$) and the neighbor masking rate ($\alpha$). Generally, \ttt{SIGNA} is relatively more sensitive to $p$ than to $\alpha$, especially when $p>0.5$. 

For $p$, we can find that as the value of $p$ grows from 0 to 0.9, the performance first improves slowly, reaching optimal performance around $p=0.4$, after which the performance decreases sharply as $p$ continues to increase. This is because too small $p$ cannot bring enough diversity for embeddings, while too large $p$ will cause embeddings to lose necessary expressive power. Therefore, we recommend a medium value for $p$ when implement SIGNA on new datasets.

While for $\alpha$, the best performance is achieved when 
$\alpha=0.4$, while both too small and too large neighbor masking probability will yield less satisfactory performance. This is consistent with the results reported in ablation study, where both NoMask and AllMask perform badly. Besides, recalling Theorem 0.4 and Collary 0.5, the value of $\alpha$ directly controls the expectation of target similarity for neighbors. Hence, a medium $\alpha$ should be beneficial for SIGNA to pursue soft neighborhood awareness.

% \section{Acknowledgments}
% AAAI is especially grateful to Peter Patel Schneider for his work in implementing the original aaai.sty file, liberally using the ideas of other style hackers, including Barbara Beeton. We also acknowledge with thanks the work of George Ferguson for his guide to using the style and BibTeX files --- which has been incorporated into this document --- and Hans Guesgen, who provided several timely modifications, as well as the many others who have, from time to time, sent in suggestions on improvements to the AAAI style. We are especially grateful to Francisco Cruz, Marc Pujol-Gonzalez, and Mico Loretan for the improvements to the Bib\TeX{} and \LaTeX{} files made in 2020.

% The preparation of the \LaTeX{} and Bib\TeX{} files that implement these instructions was supported by Schlumberger Palo Alto Research, AT\&T Bell Laboratories, Morgan Kaufmann Publishers, The Live Oak Press, LLC, and AAAI Press. Bibliography style changes were added by Sunil Issar. \verb+\+pubnote was added by J. Scott Penberthy. George Ferguson added support for printing the AAAI copyright slug. Additional changes to aaai25.sty and aaai25.bst have been made by Francisco Cruz and Marc Pujol-Gonzalez.

% \bigskip
% \noindent Thank you for reading these instructions carefully. We look forward to receiving your electronic files!

% \bibliography{aaai25.bib}
% \bibliographystyle{plainnat}

    % Second group of three images

\end{document}